\documentclass[letterpaper, 10 pt, conference]{ieeeconf}  % Comment this line out if you need a4paper

\usepackage{xcolor} % for colors

\usepackage{tabularx} % in your preamble
\usepackage{booktabs} % for nice rules
\usepackage{graphicx}
\usepackage{multirow} % for multirow cells in tables
\usepackage{booktabs} % in the preamble
 \usepackage{lipsum}
 \usepackage{svg}

 \usepackage{pifont} % for checkmarks
\usepackage{makecell} % for better line breaks in cells
\newcolumntype{Y}{>{\centering\arraybackslash}X}
\usepackage{tablefootnote}
\usepackage{threeparttable}

\definecolor{rgbcolor}{RGB}{220,50,47}   % red for RGB-trained
\definecolor{crosscolor}{RGB}{38,139,210} % blue for cross-modal-trained

\definecolor{darkgreen}{RGB}{0, 100, 0}
\newcommand{\cmark}{{\color{darkgreen}\checkmark}}
\newcommand{\xmark}{{\color{red}\ding{55}}}

\IEEEoverridecommandlockouts                              % This command is only needed if 
\usepackage{amsmath} % assumes amsmath package installed
\usepackage{amssymb}  % assumes amsmath package installed

\newcommand{\CoolName}{AnyThermal}
\newcommand{\DatasetName}{TartanRGBT}
\usepackage{cite}
\usepackage{hyperref}
\usepackage{graphicx}
\usepackage{capt-of}
\usepackage{xcolor}

\title{\LARGE \bf
    \CoolName: Towards Learning Universal Representations for \\ Thermal Perception \\
    {\color{blue} https://anythermal.github.io/}

}

\author{Parv Maheshwari$^{1}$, Jay Karhade$^{*1}$, Yogesh Chawla$^{*2}$, Isaiah Adu$^{3}$, Florian Heisen$^{4}$, Andrew Porco$^{5}$,\\ Andrew Jong$^{1}$, Yifei Liu$^{1}$, Santosh Pitla$^{2}$, Sebastian Scherer$^{1}$, Wenshan Wang$^{1}$% <-this % stops a space
\thanks{$^{*}$ Equal contribution}%
\thanks{$^{1}$ Authors are with 
Robotics Institute, Carnegie Mellon University, Pittsburgh, PA, USA. {\tt\small \{parvm, jkarhade, ajong, yifeil5, basti, wenshanw\}@andrew.cmu.edu}}%
\thanks{$^{2}$ Authors are with
Biological Systems Engineering, University of Nebraska-Lincoln, Lincoln, NE, USA. {\tt\small \{ychawla2, spitla\}@nebraska.edu}}
\thanks{$^{3}$ Authors are with Mechanical Engineering, Penn State University, University Park, PA, USA. {\tt\small ioa5099@psu.edu}}
\thanks{$^{4}$ Authors are with School of Engineering and Design, Technical University of Munich, Munich, Germany. {\tt\small florian.heisen@tum.com}}
\thanks{$^{5}$ Authors are with Mechanical Engineering, Carnegie Mellon University, Pittsburgh, PA, USA. {\tt\small aporco@andrew.cmu.edu}}%
}

\begin{document}
\makeatletter
\bstctlcite{IEEEexample:BSTcontrol}

\let\old@maketitle\@maketitle
\renewcommand{\@maketitle}{%
  \old@maketitle
  \begin{center}

  % \textbf{\CoolName} is a generalizable and task-agnostic thermal encoder that achieves strong performance across diverse tasks, including cross-modal place recognition, segmentation, and depth estimation. 
    \includegraphics[width=\linewidth]{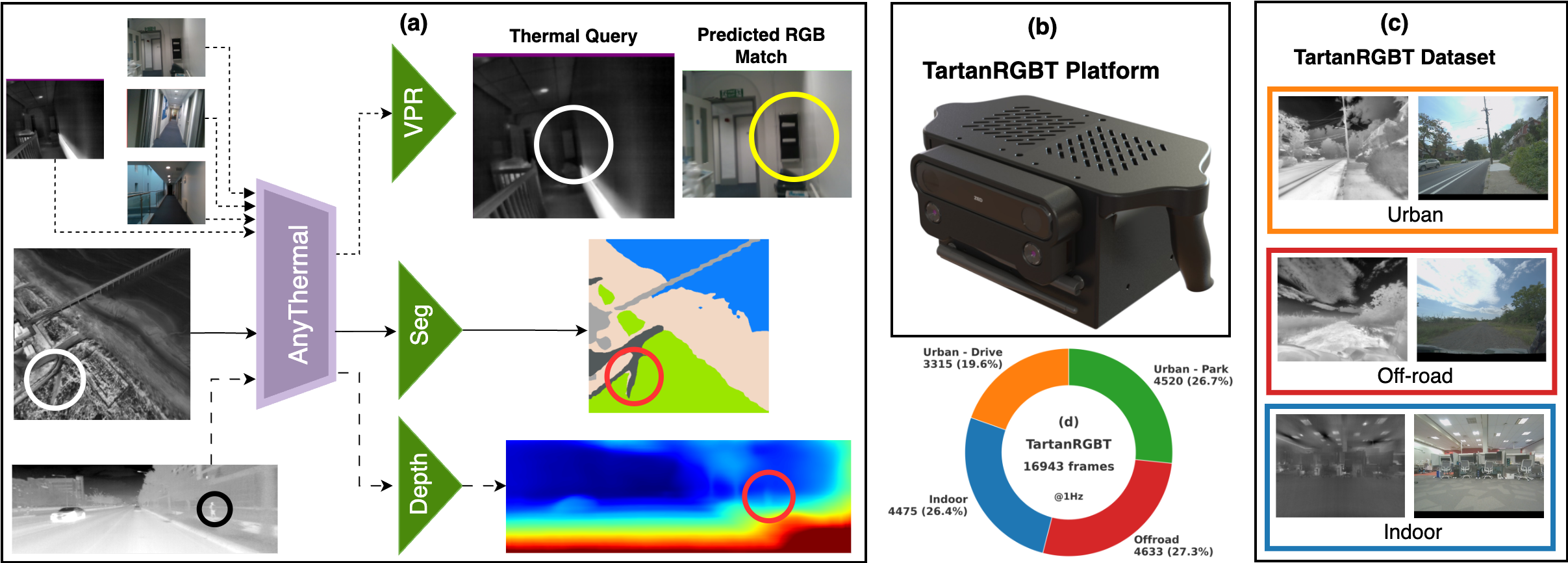}
    \captionof{figure}{\textbf{\CoolName}\ is a task-agnostic thermal encoder that delivers state-of-the-art performance across diverse tasks—such as cross-modal place recognition, thermal segmentation, and monocular thermal depth estimation—and can be applied to a wide range of environments, including indoor, aerial, off-road, and urban settings. To bridge the existing data diversity gap for training \CoolName, we build (b) an open-source data collection platform and introduce (c) \textbf{TartanRGBT}, a synchronized RGB-T dataset that spans over four types of diverse environments, as shown in (d) with a balanced distribution and a total of 16943 RGB-T pairs.}
    \label{fig:splash}
  \end{center}\par
}

% \makeatother

\maketitle

\thispagestyle{empty}
\pagestyle{empty}
  \addtocounter{figure}{-1}% <-- ensure the next figure is Fig. 2

\begin{abstract}
We present \CoolName, a thermal backbone that captures robust task-agnostic thermal features suitable for a variety of tasks such as cross-modal place recognition, thermal segmentation, and monocular depth estimation using thermal images. Existing thermal backbones that follow task-specific training from small-scale data result in utility limited to a specific environment and task. Unlike prior methods, \CoolName\ can be used for a wide range of environments (indoor, aerial, off-road, urban) and tasks, all without task-specific training. Our key insight is to distill the feature representations from visual foundation models such as DINOv2 into a thermal encoder using thermal data from these multiple environments. To bridge the diversity gap of the existing RGB-Thermal datasets, we introduce the \DatasetName\ platform, the first open-source data collection platform with synced RGB-Thermal image acquisition. We use this payload to collect the \DatasetName\ dataset - a diverse and balanced dataset collected in 4 environments. We demonstrate the efficacy of \CoolName\ and \DatasetName, achieving state-of-the-art results with improvements of up to 36\% across diverse environments and downstream tasks on existing datasets.
\end{abstract}

\section{INTRODUCTION}

The utility of thermal images has been well explored in the context of robot perception in degraded environments \cite{ms2,cart,sgm,kaist_pedestrian}. Unlike RGB sensors that are sensitive to lighting conditions and weather changes, thermal imagery is robust to all these challenges, making it a necessary addition for resilient autonomy in scenarios like search and rescue, autonomous driving, and surveillance. 

However, unlike RGB images, thermal images suffer from a scarcity of data. While RGB benefits from Internet-scale repositories that have driven major advances in deep learning \cite{dinov2,dinov3,ravi2024sam}, no such large-scale resource exists for thermal data. As a result, thermal feature extractors have yet to benefit from training at scale. Consequently, many works adapt pre-trained RGB backbones with task-specific objectives \cite{ukcheol_depth, sgm, mcnet}. In this work, we show that such RGB-only backbones fail to capture thermal-specific cues, and that using thermal images for task-agnostic training of the feature extraction backbone yields substantially stronger representations.

Since thermal datasets are scarce, a promising approach to improving thermal models is distilling knowledge from pre-trained RGB models \cite{imagebind}. This leverages both the diversity of large-scale RGB data and the correspondence between RGB and thermal views of the same scene. Effective knowledge distillation, even in data-constrained domains, requires sufficient data diversity\cite{Frank2025WhatMA}. However, prior work has been limited to a single dataset from a single environment \cite{imagebind}, restricting its generality. In this paper, we address this limitation by combining RGB-T datasets from diverse domains for distillation, and show that the resulting backbone achieves state-of-the-art performance on thermal segmentation, cross-modal place recognition, and thermal depth estimation.

While several RGB-T datasets exist, most are confined to a single type of environment (Table \ref{tab:data_distribution}). To advance knowledge distillation for thermal images, there is a clear need for RGB-T datasets spanning multiple environments. To bridge this gap, we collect a new dataset across multiple environments and demonstrate that our diverse dataset can further amplify the gains achieved from distillation.

% In this paper, we aim to overcome this limitation by distilling prior knowledge from pre-trained RGB priors into thermal encoders\cite{hinton2015distilling, imagebind} by combining RGB-T dataset from diverse domains. We show that this approach leads to state-of-the-art accuracy in multiple downstream tasks such as thermal segmentation, cross-modal place recognition, and depth estimation \todo{Parv - give numbers}. 

We summarize our main contributions as follows:

\begin{itemize}
    \item \CoolName: a task-agnostic feature extractor for thermal images obtained through knowledge distillation between RGB and thermal images. We show that \CoolName\ when combined with task-specific heads, achieves state-of-the-art performance across environments on downstream tasks like thermal segmentation and cross-modal place recognition, while outperforming RGB-based backbones of comparable size for tasks like monocular depth estimation using thermal images.
    
    \item \DatasetName\ platform: an open source data collection platform for collecting simultaneously captured stereo RGB and stereo thermal images. To the authors' best knowledge, this is the first open-source data collection platform for thermal images.
    \item \DatasetName\ dataset: we collect a diverse, balanced data set using the \DatasetName\ platform. The dataset covers residential areas, campuses, indoor environments, off-road terrain, parks, and trails. We also show how this dataset can further boost \CoolName's performance in various thermal downstream tasks across environments. 
\end{itemize}

We will release the models and code for \CoolName, and open-source the \DatasetName\ platform  along with the collected \DatasetName\ dataset upon acceptance.
\section{Related Works}
\subsection{Thermal Images for Robot Perception}
Thermal images have been applied to odometry \cite{vivid++,odom_beyond_vision,sthereo}, cross-modal place recognition \cite{sgm}, segmentation \cite{mfnet,cart}, detection \cite{kaist_pedestrian}, and depth estimation \cite{ms2,firestereo} across environments including indoor \cite{odom_beyond_vision,vivid++}, aerial \cite{sgm,cart}, off-road \cite{m2p2}, and urban \cite{kaist_pedestrian,ms2}. 
Although thermal algorithms have diverse applications, they are often studied in narrow tasks or domains, limiting their utility. In contrast, we evaluate \CoolName\ across a wide sets of tasks and environments to showcase its robustness and utility as a thermal encoder.

\subsection{Multi-modal Foundation Models}
Foundation models \cite{dinov2, dinov3, radford2021learning} have shown that large-scale pretraining enables generalized vision and language backbones. This has motivated robotics to adopt them for other modalities, with works like \cite{imagebind} demonstrating distillation of visual models into non-visual domains and building multimodal representations. Distillation has proven effective for depth/lidar \cite{liu2023segment, puy2024three, 2024-142542, xia2024uniloc}, improving tasks such as segmentation, classification, and place recognition. Success in transferring foundation model priors to the thermal domain \cite{imagebind} has been limited by the use of scarce and non-diverse datasets. With \CoolName, we show that training on multiple datasets enables effective distillation of foundation model priors, given that the datasets are collectively diverse.

\subsection{RGB-T Datasets}
\label{section:review_datasets}
Recent RGB–T datasets span urban\cite{ms2,sthereo,vivid++,heatnet}, indoor\cite{odom_beyond_vision}, aerial\cite{cart,sgm} and off-road \cite{cart,m2p2}, yet most cover only a single environment (Table~\ref{tab:data_distribution}). Moreover, each uses a distinct acquisition platform, and this non-standardization limits scalable, diverse collection.  As realistic thermal simulation is not yet feasible, research progress with thermal images relies on real-world data collection, highlighting the need for community-driven efforts to collect data across environments and embodiments. To lower this barrier, we will be open-sourcing our \DatasetName\ platform, whose efficacy is demonstrated through \DatasetName\ dataset (Section~\ref{section:results_scaling}).

\section{\CoolName: Thermal Feature-Extraction Backbone}
\begin{figure}
\centering
\vspace{0.2cm}
\includegraphics[width=0.8\linewidth]{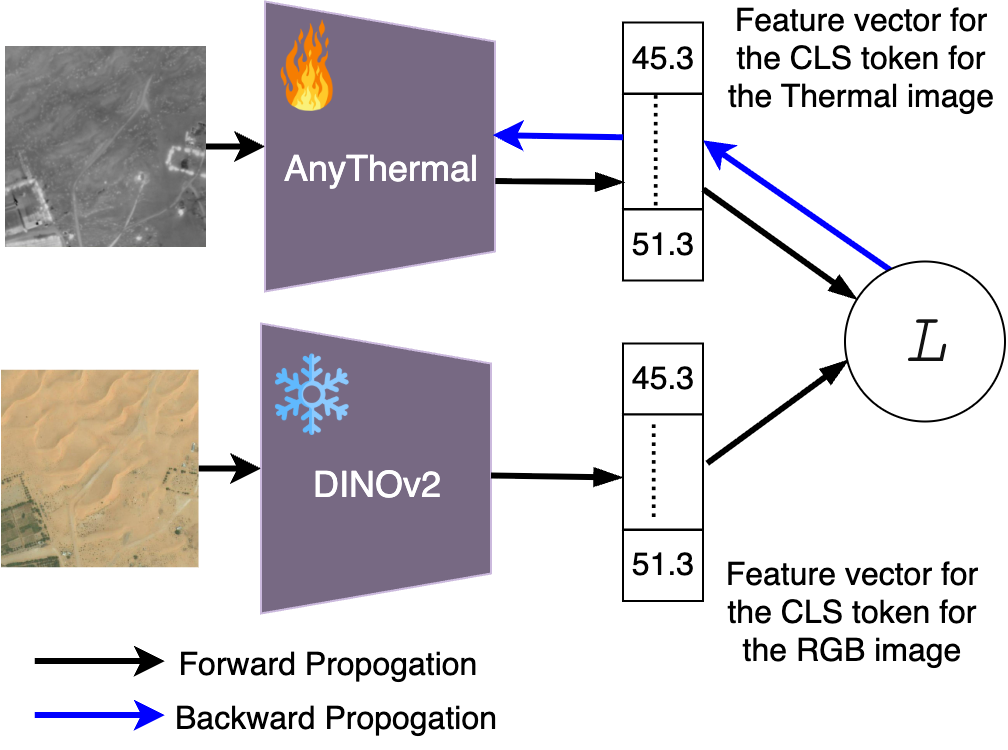}
\caption{\footnotesize We perform knowledge distillation between a frozen DINOv2 and a trainable DINOv2 network (\CoolName), both initialized with pre-trained DINOv2 weights. The frozen network serves as the teacher, while the trainable \CoolName\ backbone learns from it. Pre-trained initialization enables \CoolName\ to generalize across environments, and distillation on thermal images allows it to extract meaningful thermal features. Training is task-agnostic, using self-supervised losses between thermal features from \CoolName\ and RGB features from the frozen teacher. This approach requires no labels and scales naturally with increasing RGB-T datasets.}
\label{fig:student_teacher}
\end{figure}
\subsection{Overview}
\CoolName\ is a DINOv2-based model that has undergone knowledge distillation for thermal images. To improve generalizability across domains, the distillation is done by combining multiple datasets across domains (urban, aerial, indoor, off-road). Moreover, similar to DINOv2, we show that using \CoolName\ as a feature extraction backbone combined with a task-specific head can lead to state-of-the-art performance on tasks like thermal segmentation, cross-modal place recognition, and monocular depth estimation.
\subsection{Knowledge Distillation}

To perform knowledge distillation, two DINOv2 ViT-B/14 encoders are used. Both are initialized with pretrained weights. The teacher network processes RGB images and is kept frozen, while the student processes thermal images and is trainable (Fig. \ref{fig:student_teacher}). To use DINOv2 encoders with thermal images, the images are converted from grayscale to 3-channel. After distillation, the student serves as our \CoolName\ model.

For RGB–thermal knowledge distillation, we apply a contrastive loss on CLS token features, leveraging the intuition that corresponding RGB–thermal pairs should share similar global semantics. CLS token features from the final layer of DINOv2 capture semantic information \cite{dinov2}, rather than low-level cues like color, making them a strong basis for alignment. Moreover, using contrastive loss on the CLS token, as compared to any form of patch loss (losses calculated on corresponding patches from the RGB–thermal pair), also relaxes constraints on RGB-thermal image alignment or exact synchronization. This is particularly advantageous when distilling using datasets like VIVID++ and STheReO, where perfectly aligned RGB–thermal pairs or precise time-sync are not available.

% \[
% \mathcal{L} = - \frac{1}{B} \sum_{i=1}^B 
% \log \frac{\exp\left( \frac{ \mathrm{sim}(\mathbf{s}_i, \mathbf{t}_i)}{\tau} \right)}
% {\sum_{j=1}^N \exp\left( \frac{ \mathrm{sim}(\mathbf{s}_i, \mathbf{t}_j)}{\tau} \right)},
% \]

% where:
% \begin{itemize}
%     \item $\mathbf{s}_i \in \mathbb{R}^D$ is the student embedding for sample $i$,
%     \item $\mathbf{t}_i \in \mathbb{R}^D$ is the teacher embedding for sample $i$,
%     \item $\tau > 0$ is the temperature scaling factor,
%     \item $B$ is the batch size,
%     \item $\mathrm{sim}(\mathbf{a}, \mathbf{b})$ denotes the similarity function, typically cosine similarity.
% \end{itemize}

We used five datasets to train \CoolName, distributed as:
\begin{itemize}
    \item Urban: ViVID++ (Outdoor Driving Sequences)\cite{vivid++}, STheREo\cite{sthereo}, Freiburg\cite{heatnet} and \DatasetName\ (ours)
    \item Aerial: Boson Nighttime Dataset \cite{sgm}
    \item Indoor: \DatasetName\ (ours)
    \item Offroad: \DatasetName\ (ours)
\end{itemize}

Other datasets such as MS$^2$\cite{ms2}, CART\cite{cart}, and OdomBeyondVision\cite{odom_beyond_vision} are reserved for zero-shot evaluation on downstream tasks. M2P2, despite its large size of off-road sequences, is excluded from training \CoolName\ because many sequences have poor visibility, which weakens RGB teacher features and hampers effective thermal distillation.\begin{figure*}[t]
    \centering
    \vspace{0.2cm}
    \includegraphics[width=0.35\linewidth]{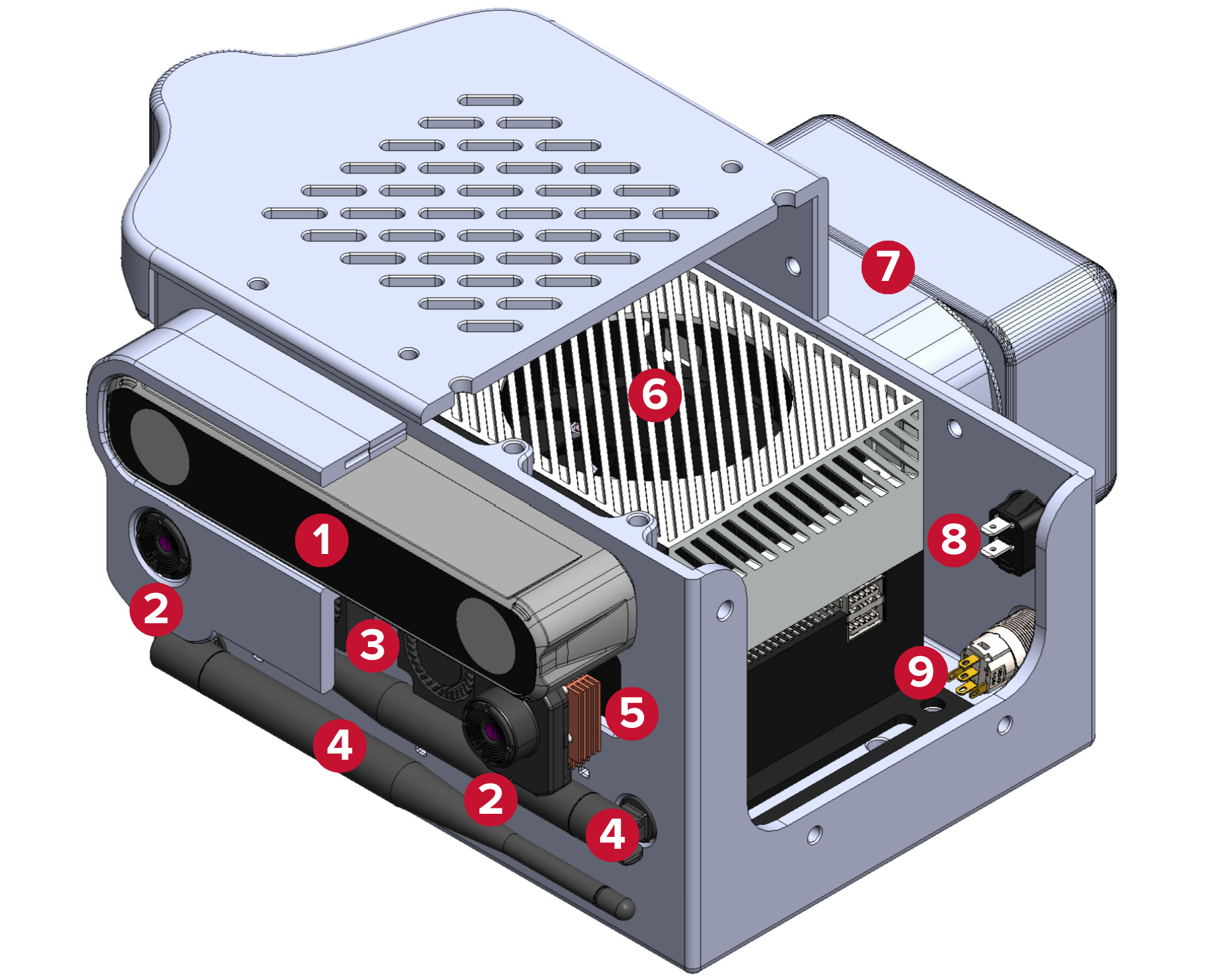}
    \includegraphics[width=0.55\linewidth]{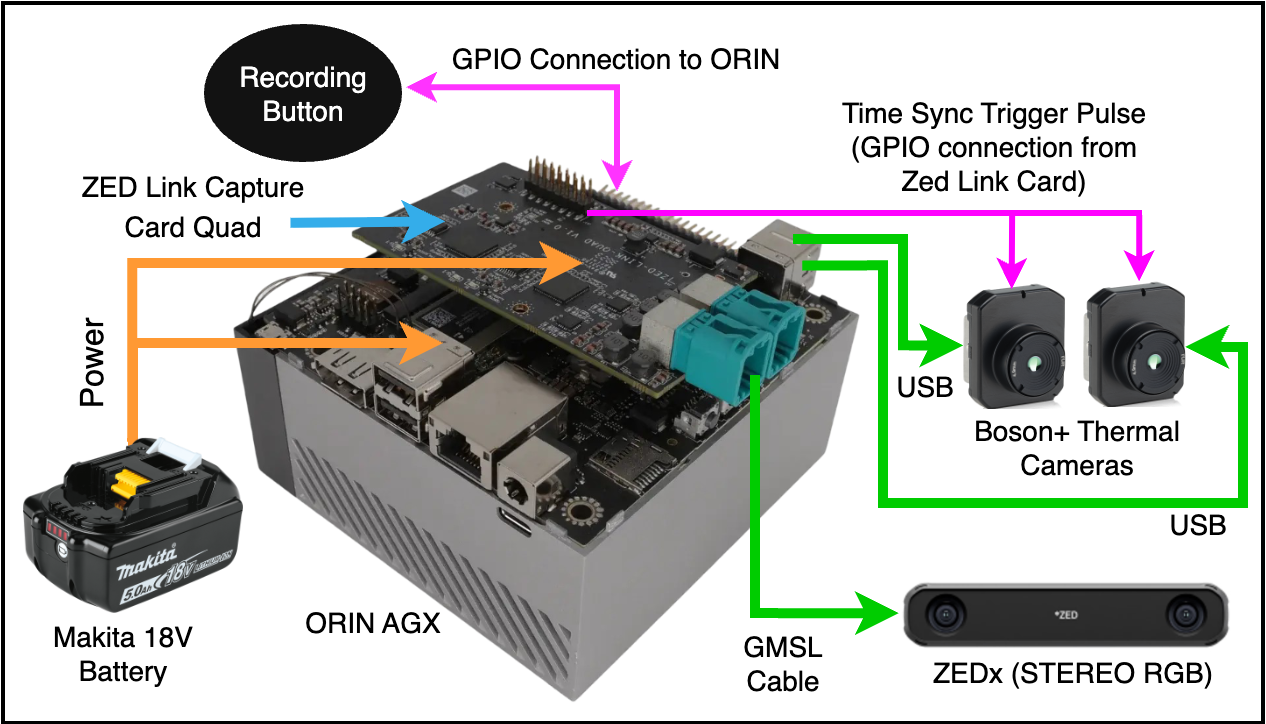}
    \caption{\textbf{Left:} CAD model of the \DatasetName\ system with half of the camera's and payload's casing hidden. Numbered components: (1) ZED X stereo camera; (2) Teledyne FLIR Boson 640 × 512, 4.9 mm, 95° HFoV, short-lens Shutterless LWIR thermal camera; (3) 5 V, 30 mm blower fan; (4) Wi-Fi antennae; (5) copper heat sinks (surrounding the thermal camera body); (6) NVIDIA Jetson AGX Orin Developer Kit, 64 GB; (7) Makita 18 V LXT® lithium-ion 4.0 Ah battery with adapter; (8) power switch; (9) recording button. \textbf{Right:} Overview of the connections between components, showing power (orange), sensor data transfer (green), and signal transfer(pink) —time synchronization and recording button trigger.}
    \label{fig:payload_cad}
\end{figure*}
\subsection{Task-Specific Head and Training}

As the feature descriptors from a ViT-based model can be quite large, they are combined with task-specific heads, which can be trained for a particular task like segmentation, visual place recognition (VPR), depth estimation, etc. In Section \ref{sec:results}, we showcase how \CoolName, when combined with task-specific heads, can lead to state-of-the-art performance on downstream tasks.
% three tasks - cross-modal place recognition, thermal segmentation, and monocular thermal depth estimation.
\subsection{Cross-Modal Place Recognition}
\label{section:methods_vpr}

A cross-modal place recognition task is to find a positive match in a database ($D$) of the modality $A$ for a query ($q$) of modality $B$. 
% Similar to \cite{sgm}, we use thermal queries and RGB databases. 
Similar to \cite{sgm}, we use thermal queries, and a corresponding RGB database.
Moreover, for each training dataset, an environment-specific radius defines ground-truth positives, chosen as a geographical radius when odometry/GPS is available or a temporal(frame) radius otherwise.

For VPR, methods like SALAD\cite{salad} and SGM\cite{sgm} show benefits of pairing a feature extractor \cite{dinov2,resnet} with a specialized head (NetVLAD\cite{netvlad}, SALAD). We choose SALAD due to its higher recall compared to other VPR heads \cite{salad}.

Following \cite{sgm}, we train with a triplet margin loss \cite{tripletloss}, where each triplet \((a,p,n)\) consists of an anchor (RGB or thermal image), a positive, and a negative. All datasets used for knowledge distillation also train the VPR head, ensuring robust clustering across environments. Unlike distillation, VPR training uses intra-dataset sampling to form harder, visually similar triplets for more effective learning.
% \paragraph{Batch Composition.}  
% A minibatch $\mathcal{B}$ is drawn from one dataset $\mathcal{D}_j \in \mathcal{D}$, each with distance function $d_j(\cdot,\cdot)$:
% \[
% \mathcal{B} = \{(x_i^{\text{rgb}},\, x_i^{\text{thr}})\}_{i=1}^B.
% \]

% \paragraph{Triplet Definition: $\forall\ x_a \in \mathcal{B}$}  
% \begin{align}
% \mathcal{P}(a) &= \{\,x_p \in \mathcal{B} \;\mid\; d_j(a,p) \le r_{+}\,\}, \\
% \mathcal{N}(a) &= \{\,x_n \in \mathcal{B} \;\mid\; d_j(a,n) \ge r_{-}\,\},
% \end{align}
% with $r_{-}>r_{+}$ to filter false negatives \cite{sgm}. Each anchor’s paired RGB–thermal image is always in $\mathcal{P}(a)$.  

% \paragraph{Mining Strategy.}  
% To mine triplets, we compute similarities and losses between the embeddings ($e_a,e_p,e_n$):  
% \begin{align}
% s_{ap} &= e_a^\top e_p, & s_{an} &= e_a^\top e_n, \\[2pt]
% \ell(a,p,n) &= \max\!\bigl(0,\, m + s_{an} - s_{ap}\bigr),
% \end{align}
% then select $\mathcal{N}$ negatives per $(a,p)$:  
% \begin{align}
% \mathcal{H}_{ap} &= \text{hardest } n_{\text{hard}} \text{ by } \ell, \\
% \mathcal{S}_{ap} &= \text{semi-hard } n_{\text{semi}} \text{ via Gumbel-Top-K}, \\
% \mathcal{R}_{ap} &= \text{hard-tail fill}, \\[4pt]
% \mathcal{T}_{ap} &= (\mathcal{H}_{ap} \cup \mathcal{S}_{ap} \cup \mathcal{R}_{ap}), \quad 
% |\mathcal{T}_{ap}| = \mathcal{N}.
% \end{align}

\subsection{Thermal Segmentation}
\label{section:method_segmentation}

As suggested in DINOv2 \cite{dinov2}, ViT feature extractors (e.g., DINOv2, \CoolName) can pair with lightweight heads for segmentation. After ablations with a single-layer MLP, two-layer non-linear MLP, and a DPT head, we select the two-layer non-linear MLP for \CoolName. It takes patch features of size $(H/14 \times W/14 \times 768)$ from DINOv2 and outputs a mask $(H/14 \times W/14 \times C)$ for $C$ classes, which is upsampled and compared with the ground truth. The backbone remains frozen, training only the head with Dice loss \cite{diceloss}, which outperformed cross-entropy. To mitigate data scarcity, we apply augmentations including brightness, contrast, gamma, and horizontal flipping.

 \subsection{Mono-Thermal Depth Estimation}
\label{section:method_depth}

For monocular depth estimation, we use the training and evaluation code for MiDaS \cite{midas} framework from \cite{ukcheol_depth}, which originally uses an EfficientLite3 backbone. We replace this with ViT-based backbones (frozen DINOv2 or \CoolName), using multiscale patch features from different layers to mimic EfficientNet3’s hierarchical features. The rest of the MiDaS architecture remains unchanged.

\section{\DatasetName\ Platform}

To collect RGB-T pairs in diverse environments, we have designed a data collection platform - \DatasetName\ platform, as shown in Fig. \ref{fig:payload_cad}, which comprises a compute module (NVIDIA Orin AGX 64GB), an 18V Makita battery, a ZEDx camera (stereo RGB + IMU), a ZEDx quad link capture card, and two FLIR Boson 640+ cameras (stereo-thermal). The sensors are hardware timesynced and capture data at 30Hz.

\subsection{CAD design and 3D printing}

The payload, as shown in Fig. \ref{fig:payload_cad}, is housed in a custom 3D-printed case with ergonomic handles on the top and sides for ease of use. Each thermal camera has heatsinks and an active cooling fan to maintain stable operation. The enclosure provides access to external ports and includes air vents to ensure airflow around the onboard computer. 

\subsection{Time Syncing}
In our platform, all four cameras (2 RGB, 2 thermal) are hardware-synchronized. The stereo RGB pair from the ZEDx is factory-synced, while a trigger pulse from the ZED Link Capture Card Quad synchronizes the thermal cameras. The pulse, aligned with RGB frame capture, is fed to both thermal cameras via their external sync pins. Configured in SLAVE mode with the BOSON SDK, the thermal cameras capture at 30 FPS in sync with the RGB cameras.

\subsection{Calibration}
\begin{figure}
\centering
\setlength{\tabcolsep}{1pt}
\begin{tabular}{cc}
    \includegraphics[width=0.4\linewidth]{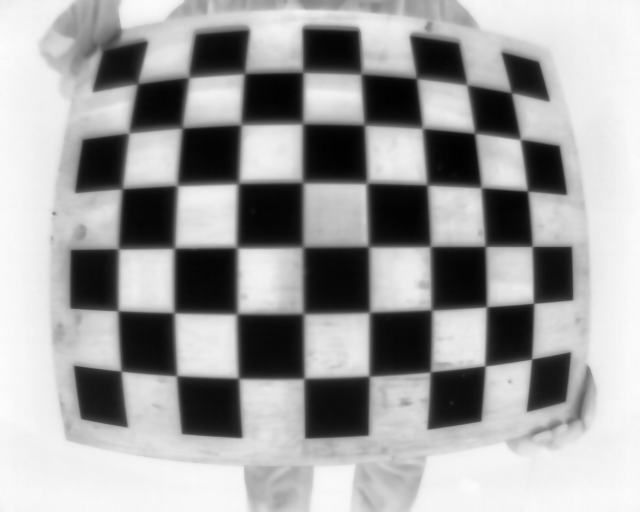} &
    \includegraphics[width=0.4\linewidth]{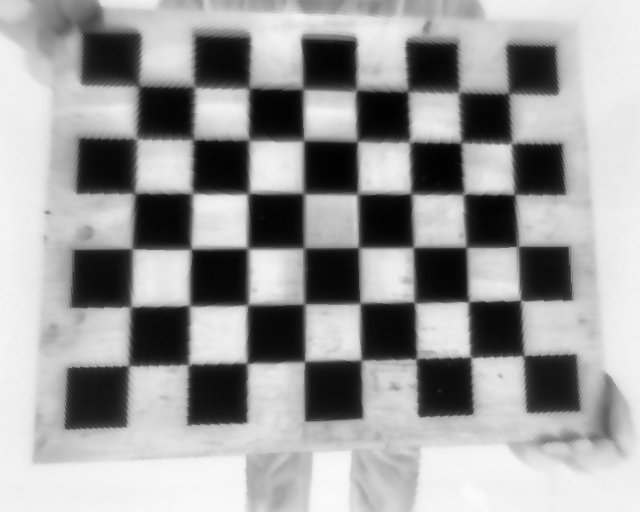}
\end{tabular}
\caption{Thermal checkerboard calibration image before (left) and after (right) fisheye rectification}
\label{fig:rectified_thermal}
\end{figure}
A complete calibration of all cameras requires intrinsic, distortion, and extrinsic factors between the cameras. Factory calibration of the stereo RGB pair was used to retrieve the intrinsics and distortion coefficients of each RGB camera. To calibrate the intrinsics and distortion parameters of the thermal cameras, a custom heated checkerboard was used similar to \cite{thermal_board}. The results after thermal rectification can be seen in Fig. \ref{fig:rectified_thermal} . The extrinsics between the RGB and thermal cameras were retrieved from the CAD design.

\subsection{Data Collection Procedure}
% For ease of data collection, the payload includes:  
% \begin{itemize}
%     \item \textbf{Bootup Procedure}: All sensor drivers (cameras, ROS2 recording, GPIO for the recording button) are launched via Docker at boot, removing the need for manual setup.  
%     \item \textbf{Recording Button}: A hardware button allows easy start/stop of recordings.  
%     \item \textbf{WiFi Antennas}: External antennas enable remote access to the ORIN.

% \end{itemize}
For ease of data collection, the payload auto-launches all sensor drivers (cameras, ROS2 recording, GPIO) via Docker at startup, eliminating manual setup. A hardware button enables one-click start/stop of recordings, and external WiFi antennas provide remote access to the ORIN.

 \subsection{Open-Source}
 \label{section:platform_opensource}
In order to open-source the \DatasetName\ platform, care has been taken to use easily available parts for assembly. Upon acceptance, we will release the CAD files, software stack (Docker, sensor drivers), component list, and assembly instructions. With this, our hope is to lower the entry barrier for the research community to collect RGB-T data. 

\begin{table*}[h]
\centering
\vspace{0.2cm}

\caption{COMPARISON OF RGB-T DATASETS ACROSS SENSING MODALITIES, SYNCHRONISATION, AND ENVIRONMENTS. 
}
\label{tab:data_distribution}

\begin{center}

\begin{threeparttable}
\begin{tabularx}{\textwidth}{|Y|c|c|c|c|c|c|ccccc|}
\hline
\multirow{2}{*}{\textbf{Dataset}} & \multirow{2}{*}{\textbf{Plat.}} & \# RGB-T Pairs &
\multirow{2}{*}{\textbf{RGB}} & \multirow{2}{*}{\textbf{THR}} & 
\multirow{2}{*}{\textbf{Sync}} & \multirow{2}{*}{\textbf{Reg.}} & 
\multicolumn{5}{c|}{\textbf{Environment}} \\
\cline{8-12}
 & &@1Hz\tnote{a} & & & & & Indoor & Offroad & Aerial & U-Drive & U-Park \\
\hline
MS$^2$\cite{ms2} & V & 16215 & \textcolor{darkgreen}{S} & \textcolor{darkgreen}{S} & \cmark & \xmark & \xmark & \xmark & \xmark & \cmark & \xmark \\
ViVID++\cite{vivid++} & H/V & 14824 & \textcolor{red}{M} & \textcolor{red}{M} & \cmark & \xmark & \xmark\tnote{b} & \xmark & \xmark & \cmark & \xmark \\
STheReO\cite{sthereo} & V & 8393 & \textcolor{darkgreen}{S} & \textcolor{darkgreen}{S}\tnote{c} & \xmark & \xmark & \xmark & \xmark & \xmark & \cmark & \xmark \\
CART\cite{cart} & H/D & 9678 & \textcolor{red}{M} & \textcolor{red}{M} & \cmark & \cmark & \xmark & \cmark & \cmark & \xmark & \xmark \\
Boson-Nighttime\cite{sgm} & D & 52590/N\tnote{d} & \textcolor{red}{M} & \textcolor{red}{M} & \xmark & \cmark & \xmark & \xmark & \cmark & \xmark & \xmark \\
OdomBeyondVision\cite{odom_beyond_vision} & D/G/H & 7129 & \textcolor{darkgreen}{S} & \textcolor{red}{M} & \xmark & \xmark & \cmark & \xmark & \xmark & \xmark & \xmark \\
M2P2\cite{m2p2} & G & 34362 & \textcolor{darkgreen}{S} & \textcolor{red}{M} & \cmark & \cmark & \xmark & \cmark & \xmark & \xmark & \cmark \\
\hline
Ours (TartanRGBT) & H & 16943 & \textcolor{darkgreen}{S} & \textcolor{darkgreen}{S} & \cmark & \cmark & \cmark & \cmark & \xmark & \cmark & \cmark \\
\hline
\end{tabularx}
\begin{tablenotes}
\item [a] Number of frames is considered at 1Hz to ensure non-redundancy of data in knowledge-distillation.
\item[b] While VIVID++ contains some indoor sequences, all of them are in a VICON cage and hence not diverse even for an indoor dataset
\item[c] The stereo thermal pair is not timesynced
\item[d] The frequency of thermal capture is not specified. So the N is unknown
\end{tablenotes}
\end{threeparttable}
\end{center}

\vspace{0.5ex}
% {\footnotesize Platform abbreviations: V = Vehicle, H = Handheld, D = Drone/UAV, G = UGV. Combinations (e.g., H/V, U/G/H) indicate multiple platforms. Reg. column indicates whether registered (aligned) RGB-T pairs are provided. Sync refers to Hardware Syncing between the RGB and Thermal cameras. U-Drive and U-Park refer to urban driving (campus, road, residential areas etc.) and park environments, respectively. As shown our dataset is the most diverse dataset while also providing synced and registered RGB-Thermal pairs.}
{\footnotesize Platform abbreviations: V = Vehicle, H = Handheld, D = Drone/UAV, G = UGV. Combinations (e.g., H/V, U/G/H) indicate multiple platforms. Reg. = registered (aligned) RGB–T pairs; Sync = hardware synchronization. U-Drive and U-Park denote urban driving (campus, road, residential areas) and park environments, respectively. As shown, our dataset is the most diverse while also providing synced and registered RGB–thermal pairs.}

\end{table*}

\section{TartanRGBT Dataset}
\subsection{Data Distribution}
% As shown in Table \ref{tab:data_distribution}, \DatasetName\ dataset is the first of its kind in diversity of environments captured while providing high-quality timesynced and registered RGB-Thermal images.
As shown in Table~\ref{tab:data_distribution}, the \DatasetName\ dataset is the first of its kind in offering broad environmental diversity alongside high-quality time-synced and registered RGB–thermal images. Although its size is moderate compared to other datasets, the emphasis on diversity during collection makes it impactful in knowledge distillation compare to existing datasets, as demonstrated in Section~\ref{section:results_scaling}.
\subsection{Modalities}

Using the \DatasetName\ platform, we record stereo RGB, stereo thermal, IMU, and thermal FFC status (manually triggered and timesynced). FFC frames are filtered since thermal capture pauses during calibration. To generate registered RGB-thermal pairs, we use FoundationStereo\cite{foundationstereo} for dense depth from stereo RGB, which will also be released. For training applications such as visual place recognition (Section~\ref{section:methods_vpr}), we generate odometry using MAC-VO\cite{macvo}.

\subsection{Thermal 8-bit Processing}

To convert 16-bit raw thermal to an 8-bit image, similar to \cite{firestereo}, we apply the following in sequence: Min-Max normalisation, CLAHE, and BilateralFilter.

\subsection{RGB-Thermal Image Registration}
\label{section:rgb_thermal_registration}

\begin{figure}[h]
\centering
\setlength{\tabcolsep}{1pt}
\begin{tabular}{ccc}
    \includegraphics[width=0.325\linewidth]{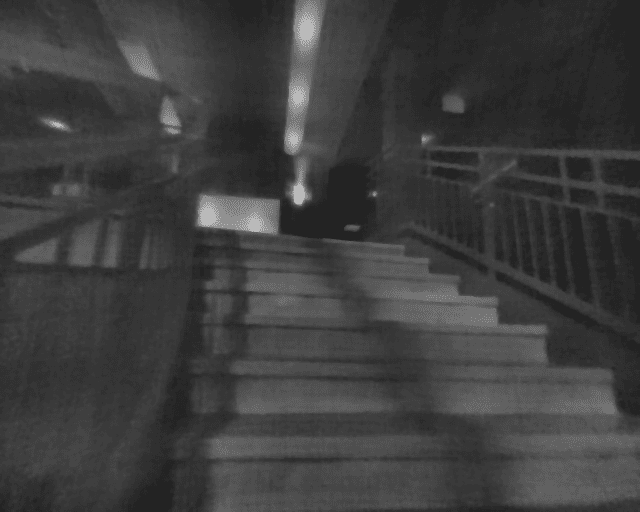} &
    \includegraphics[width=0.325\linewidth]{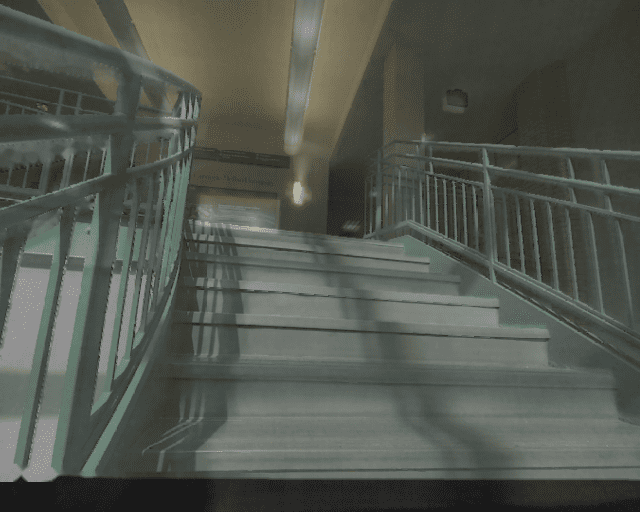} &
    \includegraphics[width=0.325\linewidth]{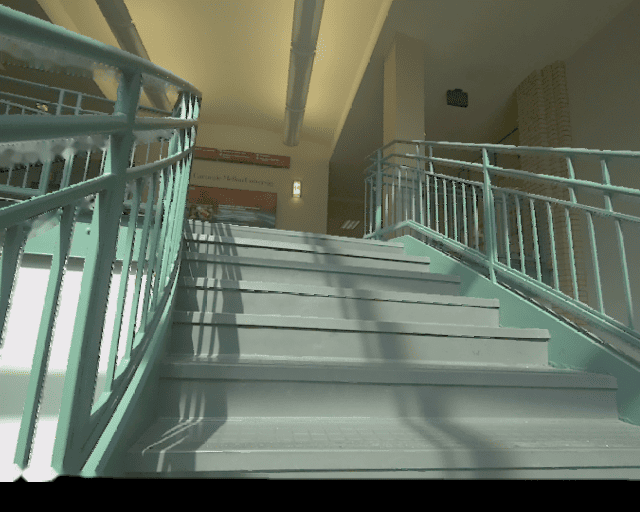} \\
    \includegraphics[width=0.325\linewidth]{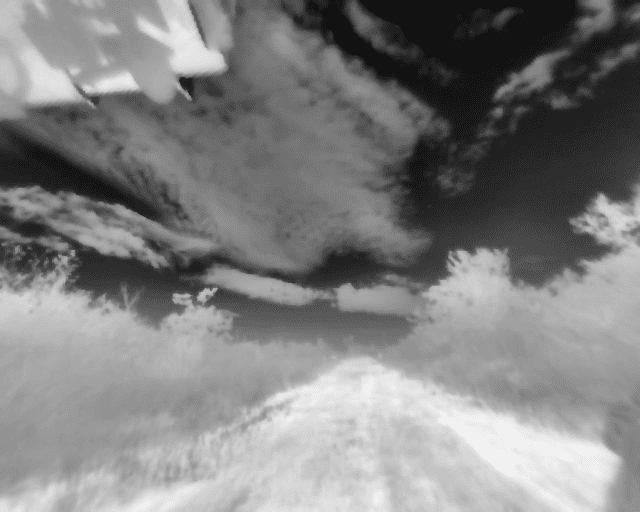} &
    \includegraphics[width=0.325\linewidth]{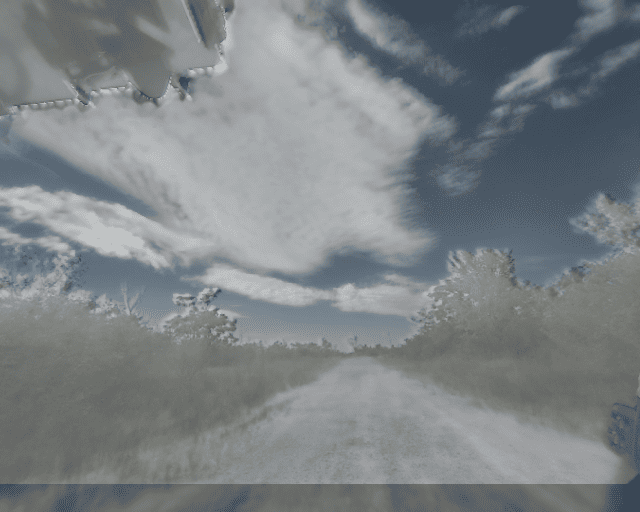} &
    \includegraphics[width=0.325\linewidth]{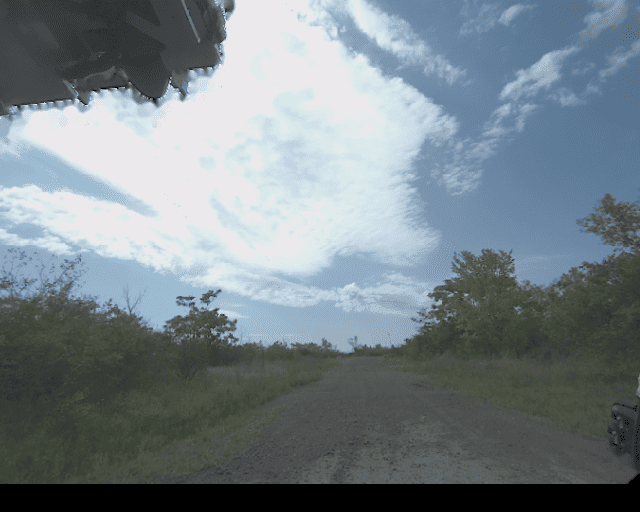} \\
    \includegraphics[width=0.325\linewidth]{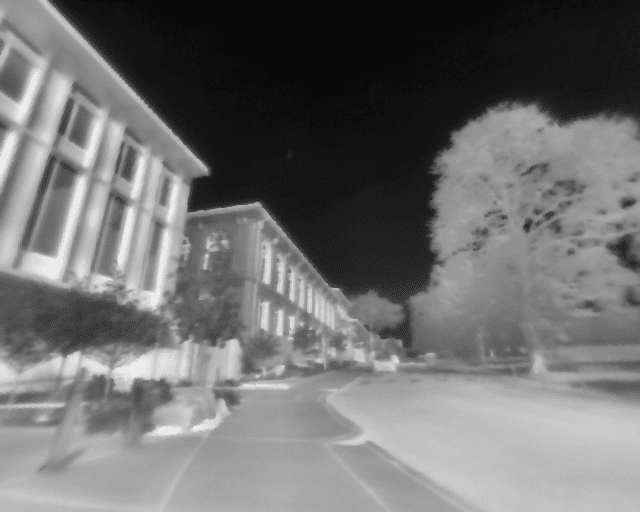} &
    \includegraphics[width=0.325\linewidth]{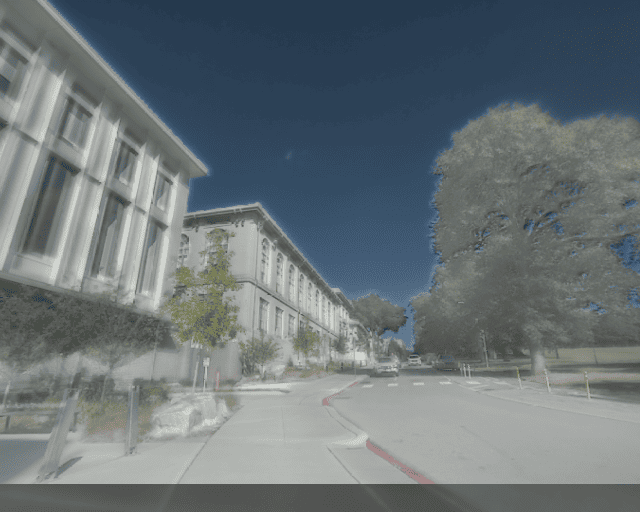} &
    \includegraphics[width=0.325\linewidth]{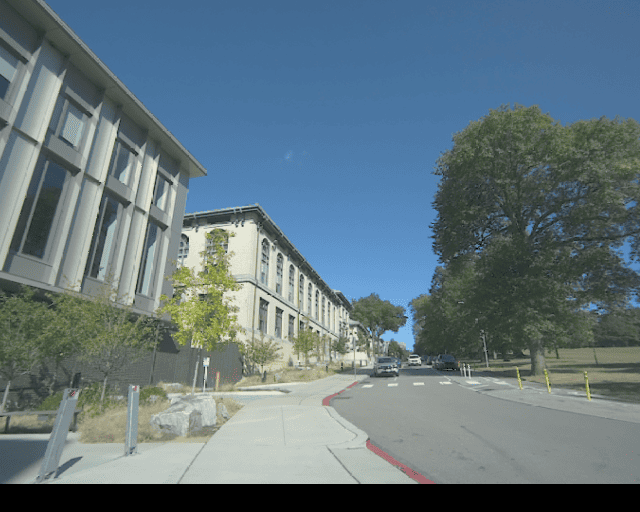} \\
\end{tabular}

\caption{\textbf{RGB–Thermal Registration} in the \DatasetName\ dataset: alpha-blended overlays for indoor, off-road, and urban domains with blending factors $\alpha \in \{0.00, 0.50, 1.00\}$. Due to sensor geometry (thermal mounted below RGB), the thermal view includes more of the lower scene, resulting in additional pixels at the bottom of the thermal images that are not present in the RGB images, producing black regions where RGB pixels are absent.}

\label{fig:rgb_thermal_alpha_sweep}
\end{figure}
Pixel-level RGB--thermal registration ensures spatial correspondence, improving distillation supervision and enabling tasks such as RGB--thermal translation, label transfer, and cross-modal learning. Existing aligned datasets (\cite{cart}, \cite{heatnet}, \cite{mfnet}) are limited and in singular environments, making our diverse dataset valuable for training and benchmarking.  

Following \cite{heatnet}, alignment has three stages: (1) estimate depth from rectified stereo RGB using FoundationStereo \cite{foundationstereo} to back-project pixels into 3D; (2) transform 3D points into the thermal frame with pre-calibrated extrinsics; (3) project with thermal intrinsics to yield aligned RGB--thermal pairs (Fig.~\ref{fig:rgb_thermal_alpha_sweep}).

% \begin{equation}
% p_{\text{RGB}}' = \phi \!\left( K_T \, T_{\text{RGB} \rightarrow T} \, \gamma(p_{\text{RGB}} \mid K_{\text{RGB}}, D_{\text{RGB}}) \right),
% \label{eq:rgb_to_thermal}
% \end{equation}

% where $p_{\text{RGB}}$ is an RGB pixel, $D_{\text{RGB}}$ is the depth estimated from FoundationStereo, $K_{\text{RGB}}, K_T$ are intrinsics, $T_{\text{RGB} \rightarrow T}$ the extrinsics, $\gamma(\cdot)$ the back-projection, $\phi(\cdot)$ the projection, and $p_{\text{RGB}}'$ the corresponding pixel in the thermal image. 

Similar to \cite{heatnet}, which employed the state-of-the-art stereo model of its time for dense depth estimation, we adopt FoundationStereo to obtain dense pixel-level alignment. Although the estimated depth is not perfect and errors in prediction directly affect the aligned outputs, it offers a practical alternative to accurate but sparse LiDAR, as knowledge distillation requires dense supervision. Furthermore, continued advances in depth estimation models are expected to further improve alignment quality.

% FoundationStereo produces a dense depth map, but during RGB-to-thermal alignment, black points were observed due to occlusions between the RGB and thermal views, as well as from rasterizing 3D points to discrete thermal pixels, which leaves some locations unfilled. To address this two steps are taken. First, a z-buffer enforces visibility by retaining only the nearest depth per thermal pixel. Second, after reprojection to 2D, bilinear splatting improves coverage by spreading each projected sample to its four neighboring pixels with interpolation weights. As shown in Fig.~\ref{fig:rgb_thermal_alpha_sweep}, in the bottom regions of the thermal images where no RGB depth is available, we do not apply splatting, since this would hallucinate content without valid 3D data.

FoundationStereo produces a dense depth map, but during RGB–thermal alignment black pixels arise from occlusions between the two views and from rasterizing 3D points onto discrete thermal pixels, leaving some locations unfilled. We address this with two steps. First, a z-buffer enforces visibility by retaining only the nearest depth per thermal pixel. Second, after projection to 2D, bilinear splatting improves coverage by distributing each projected sample across its four neighboring pixels with interpolation weights. As shown in Fig.~\ref{fig:rgb_thermal_alpha_sweep}, splatting is not applied in the lower regions of the thermal images where no RGB depth is available, as this would otherwise hallucinate content without valid 3D data.

\subsection{Limitations}

We will release dense depth and odometry to support RGB–thermal alignment and VPR training. As they are obtained from stereo-RGB algorithms, their accuracy is insufficient for benchmarking tasks such as odometry or depth estimation. Thus, we also do not evaluate downstream tasks like cross-modal place recognition or depth estimation on \DatasetName. Since VPR training does not require precise odometry, the current estimates suffice. Future work will include GPS and LiDAR for accurate odometry and depth.

\section{Results}
\label{sec:results}
We demonstrate the effectiveness of \CoolName\ on three tasks: cross-modal place recognition, thermal segmentation, and monocular thermal depth estimation.

\subsection{Cross-Modal Place Recognition}
\label{section:results_vpr}
\subsubsection{Formulation}
Our cross-modal place recognition task, as described in Section~\ref{section:methods_vpr}, is defined as: given a thermal query image, retrieve a matching RGB image from a database. To ensure proper evaluation, the paired RGB image of a query is excluded from its positive set. We report Recall@1 (R@1) in Table~\ref{tab:vpr}, where R@1 is the probability that the top retrieved match is positive for a query.

\subsubsection{Evaluation Datasets}
We evaluate \CoolName\ and baselines on three diverse zero-shot datasets: CART~\cite{cart} (aerial), MS2~\cite{ms2} (urban), and OdomBeyondVision~\cite{odom_beyond_vision} (indoor). CART and MS2 provide GPS, enabling all sequences to form a shared database, while OdomBeyondVision relies on intra-sequence odometry and is evaluated per sequence. For OdomBeyondVision, a weighted mean recall is reported across sequences, weighted by the number of queries in each.

\subsubsection{Baselines}
We compare against two categories:
\begin{itemize}
    \item \textit{RGB Methods:} R2former\cite{R2former}, NetVLAD\cite{netvlad}, MixVPR\cite{mixvpr}, and SALAD\cite{salad}. Since SALAD consistently outperforms the others, we report it as the representative RGB baseline. We also include frozen RGB-DINOv2 (teacher) without a VPR head.
    \item \textit{RGB-Thermal Methods:} ImageBind~\cite{imagebind} and SGM~\cite{sgm}. Although ImageBind is not trained for VPR, we include it since it is the only other method to perform knowledge distillation between RGB and thermal. SGM is trained for cross-modal place recognition, but only on Boson Nightime\cite{sgm}, which is an aerial-only dataset.
\end{itemize}

\begin{table}[h]
\centering
% \vspace{0.2cm}
\begin{center}
\caption{CROSS-MODAL PLACE RECOGNITION ACROSS DIVERSE ENVIRONMENTS.}
\label{tab:vpr}

\begin{threeparttable}
\begin{tabular}{c c c |c|c|c}
        \toprule
         Model Name & Backbone & Head\tnote{a}& MS$^2$& CART& OBV\tnote{b}\\
 & & & $r$: 15& $r$: 15&$r$: 3\\
        \midrule
         *DINOv2\cite{dinov2} & \textcolor{rgbcolor}{DINOv2} & X& 27.21 & 25.98& 29.49\\
          *SALAD\cite{salad}   & \textcolor{rgbcolor}{DINOv2} & \textcolor{rgbcolor}{S}& 76.97 & 49.38& 38.94\\
        \midrule
         *ImageBind\cite{imagebind} & \textcolor{crosscolor}{ViT-Huge} & X& 0.79 & 1.13& 10.25\\
             *SGM\cite{sgm}             & \textcolor{rgbcolor}{ResNet-18} & \textcolor{crosscolor}{N}&20.02 & 45.59  & 21.05\\
                 \CoolName\                     & \textcolor{crosscolor}\CoolName\    & {X}& 75.39 & 45.45 & 45.40\\
             \CoolName-VPR                 & \textcolor{crosscolor}{\CoolName}    & \textcolor{crosscolor}{S}& \textbf{81.11} & \textbf{56.00}&\textbf{53.17}\\
        \bottomrule
    \end{tabular}

\begin{tablenotes}
{ \footnotesize
\item[a] VPR Heads: \textbf{N:} NetVLAD, \textbf{S:} SALAD, \textbf{X:} No head has been used, and instead the CLS token is used as the feature vector for the images
\item[b] OBV: OdomBeyondVision\cite{odom_beyond_vision}
}
\end{tablenotes}
\end{threeparttable}
\end{center}
\vspace{0.5ex}
    % {\footnotesize The radius (\(r\), in meters) to define true matches, is chosen per environment. * denotes frozen models (backbone + head from original papers). \textcolor{rgbcolor}{Red} refers to RGB-only training, while \textcolor{crosscolor}{blue} denotes RGB--thermal training. \CoolName\ falls under the blue category, as it is initialized with RGB-pretrained weights and distilled on thermal images. The upper section lists RGB-only methods, and the lower section lists RGB--thermal methods. \CoolName, particularly with a VPR head, outperforms baselines; the gap between RGB-DINOv2 and \CoolName\ demonstrates the benefit of thermal-specific distillation of RGB-pretrained backbones.}
    {\footnotesize The positive radius (\(r\), in meters) used for determining positive matches, is chosen per environment (MS$^2$: Urban, CART: Aerial, OBV: Indoor). * denotes frozen models (backbone + head from original papers). \textcolor{rgbcolor}{Red} indicates RGB-only training, while \textcolor{crosscolor}{blue} indicates RGB--thermal training. \CoolName\ belongs to the blue category, as it is initialized with RGB-pretrained weights and distilled on thermal images. The upper section lists RGB-only methods, and the lower section lists RGB--thermal methods. \CoolName, especially with a VPR head, outperforms baselines; The gap between *DINOv2 and \CoolName\ shows the benefit of distilling RGB-pretrained backbones on thermal data.}

\end{table}

\begin{figure}
\centering
\vspace{0.2cm}
\includegraphics[width=\linewidth]{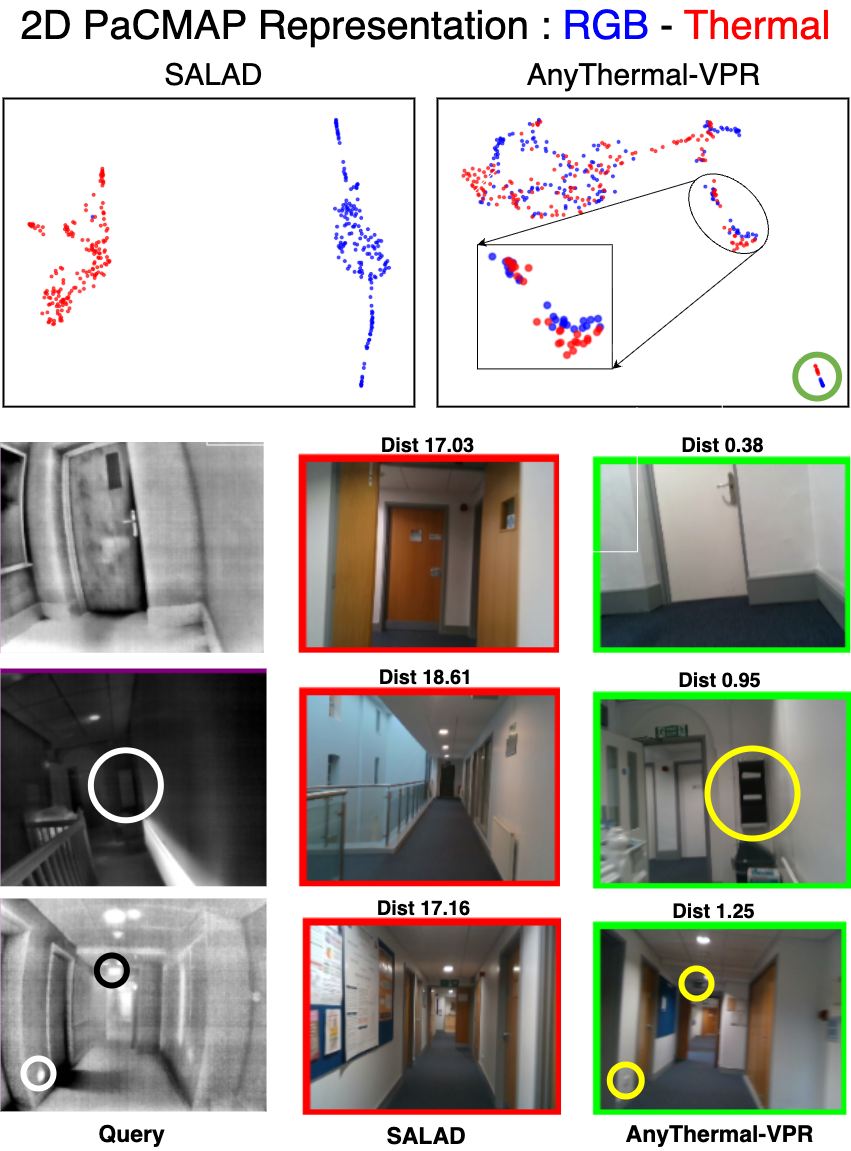}
\caption{\footnotesize \textbf{Cross-Modal VPR on OdomBeyondVision:} \textbf{Top}: PaCMAP\cite{pacmap} representations show SALAD poorly(far) aligns RGB–Thermal embeddings, while \CoolName-VPR aligns them well in a shared representation space. \textbf{Bottom}: Example queries where SALAD fails to retrieve the correct RGB match, but \CoolName-VPR succeeds, with key clues circled.
}
\label{fig:vpr_qualitative}
\end{figure}
\begin{figure*}[]
\centering
\vspace{0.2cm}
\includegraphics[width=0.7\linewidth]{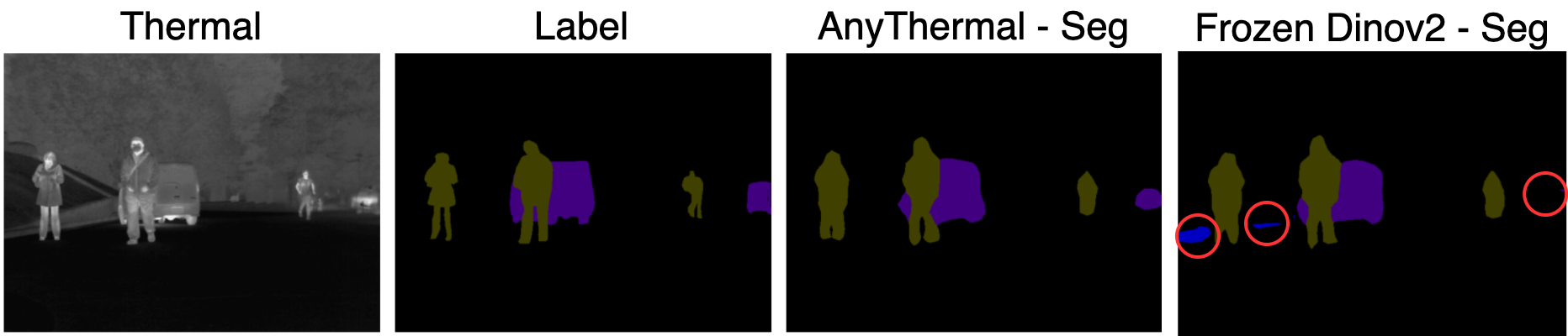}\\
\caption{\footnotesize \textbf{Thermal Segmentation on MF-Net\cite{mfnet}:} The frozen DINOv2 baseline misses objects (e.g., the car on the right) and misclassifies the background, while our AnyThermal backbone segments accurately.} 
\label{fig:qualitative_seg}
\end{figure*}

As shown in Table~\ref{tab:vpr}, \CoolName-VPR outperforms all baselines across environments. Moreover, the gap between DINOv2-X and \CoolName-X underscores the need for knowledge distillation in thermal images and confirms that frozen RGB extractors are suboptimal. This is further evidenced by \CoolName\ matching SGM’s aerial performance without a VPR head, despite both being trained solely on the Boson Nighttime dataset for aerial data. Fig. ~\ref{fig:vpr_qualitative} further illustrates that \CoolName-VPR aligns RGB and thermal representations more effectively than the strongest baseline.
\subsection{Thermal Segmentation}
\begin{table}
    \centering
    \caption{THERMAL SEGMENTATION ON MF-NET DATASET:} 
\begin{center}
    \begin{tabular}{|c|c|c|c|}
        \hline
         \textbf{Model}& \textbf{\# parameters(M)} & \textbf{mIoU (\%)} & \textbf{FPS}\\
         \hline
         RTFNET-152 \cite{rtfnet}&  196.37&  47.00\%& 8.37\\\hline
         MCNET \cite{mcnet}&  54.65 &  51.95& 1.88\\\hline
         RGB\_DINO-SEG&  87.02&  45.46\%  & 6.79\\\hline
         \CoolName-SEG&  87.02&  \textbf{53.47\%}& 6.79\\\hline
    \end{tabular}
    \end{center}
    \label{tab:segmentation}

    \vspace{0.5ex}
    {\footnotesize 
    The number of parameters is reported in Millions (M). The FPS is reported on ORIN AGX 64GB. We can see, \CoolName\ with a 2-layer MLP head (SEG) achieves state-of-the-art performance while being 3.6x faster than the closest performing baseline
    }
\end{table}

We evaluated the use of \CoolName\ for thermal segmentation (Fig. \ref{fig:qualitative_seg}) on the MF-Net\cite{mfnet} dataset using its standard train/val/test splits and all 9 classes (including background) for mIoU. Table~\ref{tab:segmentation} also reports FPS on an NVIDIA ORIN AGX 64GB. \CoolName\ achieves state-of-the-art mIoU while delivering a 3.6$\times$ FPS boost over the closest baseline.

\subsection{Mono-Thermal Depth Estimation}
\begin{table}[t]
\centering
\caption{MONOCULAR DEPTH ESTIMATION ON THE MS$^2$ DATASET}
\begin{center}
    \begin{tabular}{c|cccc} 
    \toprule
     Backbone&AbsRel$\downarrow$ & SqRel$\downarrow$ & RMSE$\downarrow$ & RMSElog$\downarrow$ \\ 
    \toprule
    efficientnet lite3 & 0.1015 & 0.3955 & 2.9587 & 0.1417 \\
    dinov2\_vitb14 & 0.0905 & 0.3177 & 2.7493 & 0.1208 \\
    \CoolName & \textbf{0.0883} & \textbf{0.3142} & \textbf{2.7432} & \textbf{0.1182} \\
    \bottomrule
    \end{tabular} 
\end{center}
\label{tab:depth_result}
\vspace{0.5ex}
{\footnotesize
We evaluate our proposed method with a representative MDE network (MiDaS~\cite{midas}). 
All results are averaged over all day, night, and rainy evaluation sets of MS$^2$.
The best performance is highlighted in \textbf{bold}.
}
\end{table}
Following \cite{ukcheol_depth}, we evaluate on the MS$^2$ dataset using sparse LiDAR ground truth and report multiple metrics (Table~\ref{tab:depth_result}). We use the MIDAS\cite{midas} architecture, where we ablate the effect of replacing the EfficientLite3 backbone used in \cite{ukcheol_depth} with frozen DINOv2, and further with \CoolName. The gain from EfficientLite3 to DINOv2 reflects network depth, while the additional improvement with \CoolName\ proves its benefits over frozen-RGB pretrained backbones.

\subsection{Scaling Data in \CoolName\ training}
\label{section:results_scaling}
\begin{figure}[]
\centering
% \includesvg[width=\linewidth]{figures/scaling_results/vpr.svg}
\includegraphics[width=\linewidth]{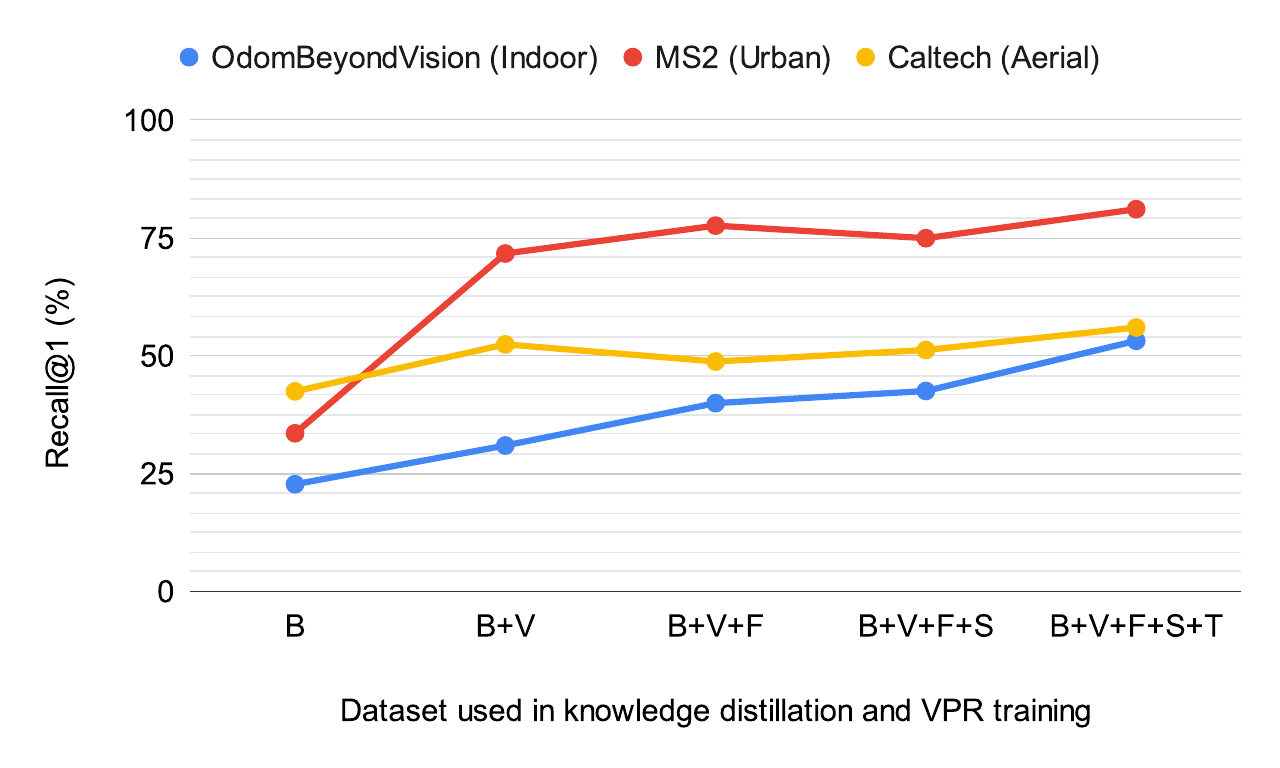}
\includegraphics[width=\linewidth]{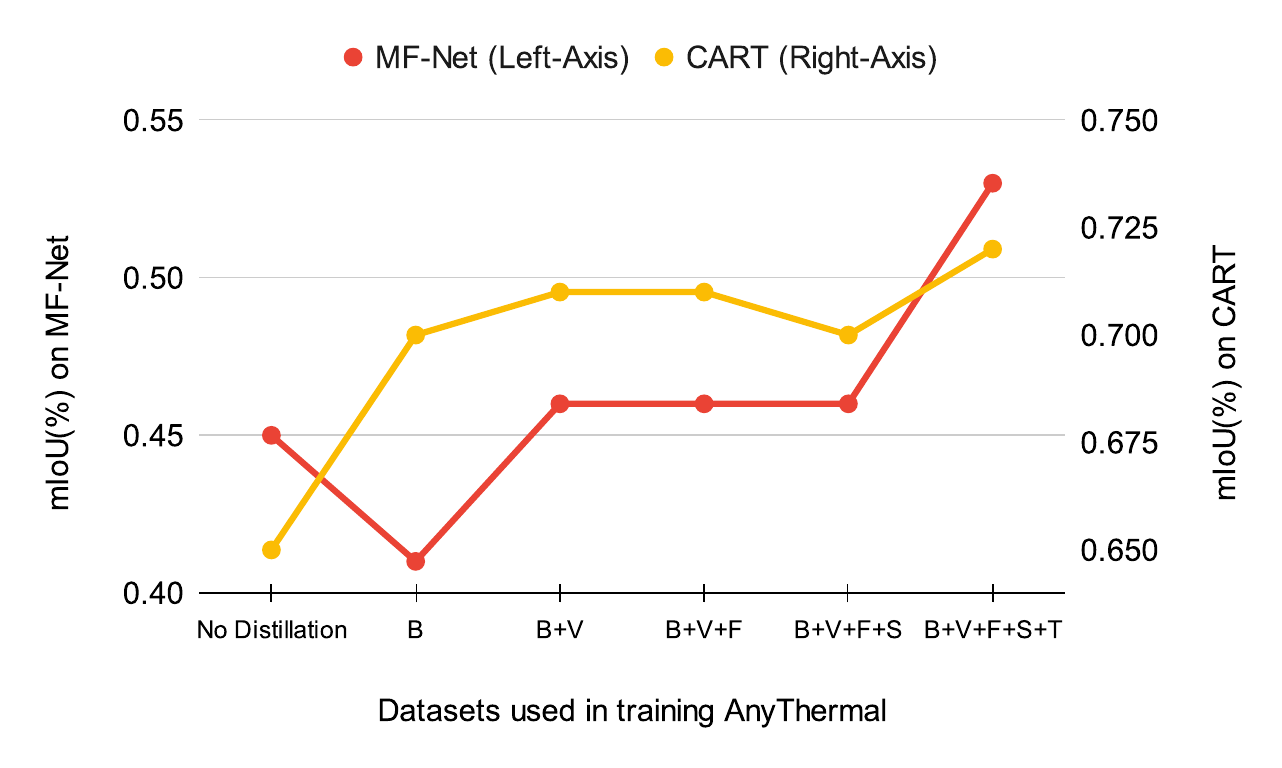}
\includegraphics[width=\linewidth]{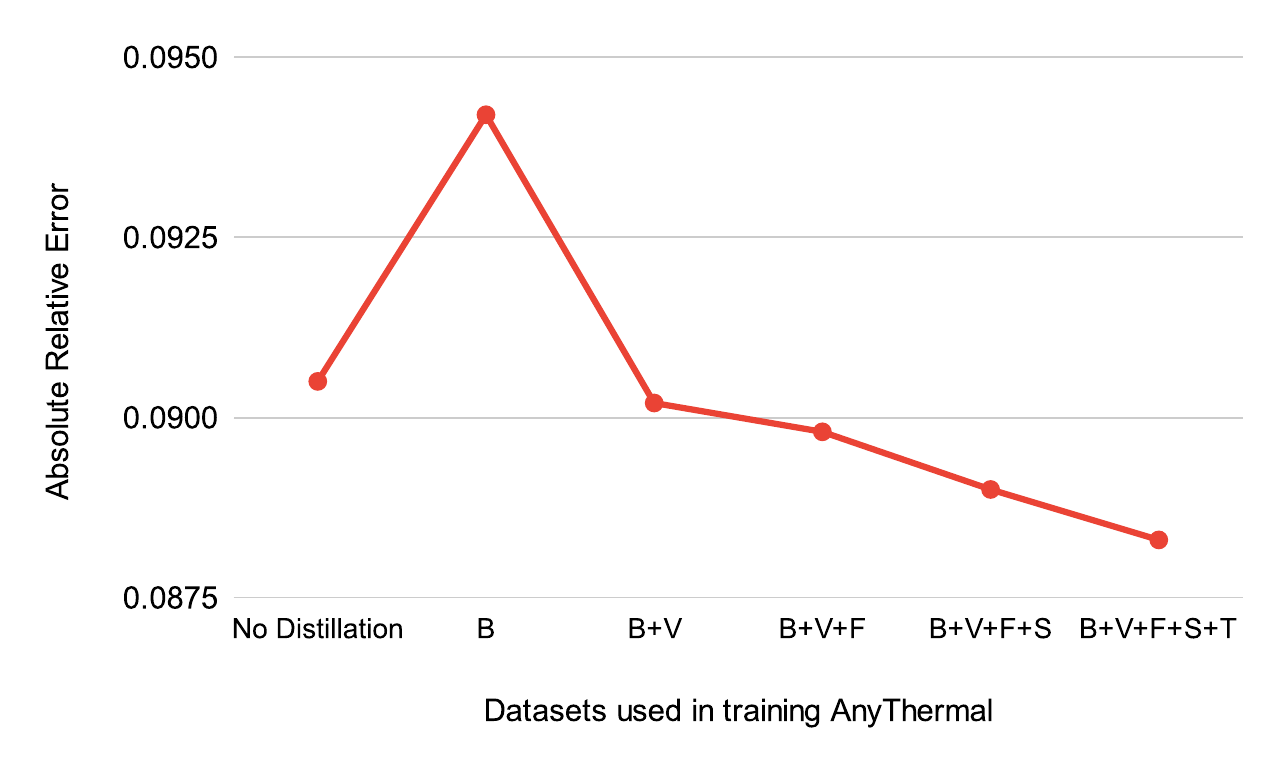}
\caption{\footnotesize Effect of scaling data in pretraining - knowledge distillation + VPR training(for top plot only)- on downstream performance. X-axis shows pretraining datasets (B: Boson Nightime, V: ViVID++, F: Freiburg, S: STheReO, T: \DatasetName). [Top]: Recall for cross-modal place recognition (higher is better). [Middle]: mIoU for thermal segmentation on MF-Net and CART (higher is better). [Bottom]: Absolute relative (Abs\_Rel) error for monocular thermal depth estimation (lower is better). Adding \DatasetName\ consistently improves performance across environments and tasks, unlike Freiburg and STheReO, which add little diversity and lead to saturation.} 
% \caption{\footnotesize Effect of data scaling in pretraining (knowledge distillation + VPR training) on downstream tasks. X-axis shows pretraining datasets (B: Boson Nightime, V: ViVID++, F: Freiburg, S: STheReO, T: \DatasetName). [Top]: Recall for cross-modal place recognition (↑). [Middle]: mIoU for thermal segmentation on MF-Net and CART (↑). [Bottom]: Abs\_Rel error for monocular thermal depth estimation (↓). Adding \DatasetName\ consistently boosts performance across environments and tasks, unlike Freiburg and STheReO, which add little diversity and saturate performance.}

\label{fig:scaling_plots}
\end{figure}

It is crucial to understand how multi-domain datasets in knowledge distillation affect downstream performance. Specifically, we ask whether simply adding more data improves efficacy, or if dataset diversity is essential for building robust feature extraction backbones.  

% We study this by scaling data during pre-training and evaluating downstream tasks. The setup involves the distillation of knowledge of the \CoolName\ backbone and VPR head training. Among task-specific heads, only VPR is included in pre-training, as it can be trained using GPS/odometry or temporal cues and evaluated zero-shot. In contrast, segmentation and depth require specialized labels, so for these tasks, evaluation and training are done on the test and train data splits from the evaluation datasets. This ensures fairness against baselines (VPR baselines are evaluated zero-shot; segmentation and depth baselines are trained on their respective datasets).  

We study the effect of data scaling during pre-training by distilling knowledge into the \CoolName\ backbone and training the VPR head. Among task-specific heads, only VPR is included in pre-training, since it can leverage GPS/odometry or temporal cues and be evaluated zero-shot. In contrast, segmentation and depth require labeled data, so these tasks are trained and evaluated on the respective splits of their evaluation datasets. This setup ensures fairness: VPR baselines are evaluated zero-shot, while segmentation and depth baselines are trained on the evaluation datasets.

As shown in Fig.~\ref{fig:scaling_plots}, adding more datasets generally improves performance but not always:
\begin{itemize}
    \item \textbf{Domain Gap in Single-Dataset Distillation:} In Fig.~\ref{fig:scaling_plots} (middle, bottom), a \CoolName\ variant distilled only on Boson Nighttime (aerial) underperforms in urban domains (red), compared to the frozen RGB-DINOv2 (No distillation). This gap arises from its aerial-only training. Conversely, performance improves on CART (middle, yellow), as it is also aerial.
    \item \textbf{Performance Saturation:} In Fig.~\ref{fig:scaling_plots}, adding more urban data (B+V $\rightarrow$ B+V+F $\rightarrow$ B+V+F+S) yields only marginal gains, with some aerial evaluations showing drops (e.g., thermal segmentation dip between $B+V+F$ and $B+V+F+S$).

\end{itemize}

In contrast, adding our \DatasetName\ consistently improves performance across tasks and domains, with notable gains in indoor VPR recall (from rich indoor sequences), improved segmentation on CART (from off-road coverage since CART segmentation includes off-road data), and even boosts in urban domains despite existing urban datasets.

These results show that while scaling data helps up to a point, data diversity is more critical than scale for building robust, generalizable feature extractors.  

\section{Conclusion and Future Work}
\label{sec:conclusion}

We present \CoolName, a task-agnostic thermal feature extraction backbone distilled from pre-trained RGB backbones. To further advance thermal research, we introduced the \DatasetName\ Platform—the first open-source RGB-T collection framework—and curated a diverse \DatasetName\ dataset. Together, \CoolName\ and \DatasetName\ deliver up to 36\% improvement across environments (urban, indoor, aerial, off-road) and tasks (cross-modal place recognition, thermal segmentation, depth estimation).

% As future work, we A) aim to apply \CoolName\ to more diverse tasks like object detection and cross modal matching, and B) distill stronger backbones leveraging newer visual foundation models\cite{dinov3}. As shown in Fig.~\ref{fig:scaling_plots}, performance has not yet plateaued, suggesting further gains through scaling. Future efforts will focus on (i) expanding \DatasetName\ with additional sensors and environments (e.g., GPS, aerial); and (ii) fostering community-driven data collection with \DatasetName\ platform to advance generalization of thermal and cross-modal algorithms.

Future directions can include A) applying \CoolName\ to more diverse tasks such as object detection and cross-modal matching, and B) distilling stronger backbones leveraging newer visual foundation models \cite{dinov3}. As shown in Fig.~\ref{fig:scaling_plots}, \CoolName's performance has not yet plateaued, suggesting further gains through scaling diverse RGB-T data. Future efforts will focus on (i) expanding \DatasetName\ with additional sensors and environments (e.g., GPS, aerial); and (ii) community-driven data collection with our platform to advance generalization of thermal and cross-modal algorithms.

% In this paper, we presented \CoolName, a general thermal encoder bakcbone that is distilled from pre-trained RGB backbones as well as introduced the \DatasetName\ Platform—the first open-source RGB-T collection framework—along with a curated and diverse \DatasetName\ dataset for advancing thermal research. Together, \CoolName\ and \DatasetName\ deliver up to 36\% improvement across environments (urban, indoor, aerial, off-road) and tasks (cross-modal place recognition, thermal segmentation, depth estimation). As future work, we A) aim to apply \CoolName to more diverse tasks like object detection and cross modal matching, and B) distill stronger backbones leveraging newer VFMs like DINOv3~\cite{dinov3}). As shown in Fig.~\ref{fig:scaling_plots}, performance has not yet plateaued, suggesting further gains through scaling. Future efforts will focus on (i) expanding \DatasetName\ with additional sensors and environments (e.g., GPS, aerial); and (ii) fostering community-driven data collection with \DatasetName\ platform to advance generalization of thermal and cross-modal algorithms.

\bibliography{IEEEabrv}

\end{document}